\pdfoutput=1

\documentclass[11pt]{article}

\usepackage[]{acl}

\usepackage{times}
\usepackage{latexsym}
\usepackage{graphicx}
\usepackage{enumitem}
\usepackage{amsmath}
\usepackage{multirow}
\usepackage{booktabs}
\usepackage{float}
\usepackage{placeins}
\usepackage{textcomp}
\usepackage [autostyle, english = american]{csquotes}
\MakeOuterQuote{"}
\usepackage{tikz}
\def\checkmark{\tikz\fill[scale=0.3](0,.35) -- (.25,0) -- (1,.7) -- (.25,.15) -- cycle;}
\usepackage{multirow}
\usepackage[misc]{ifsym}
\usepackage{subfig}

\usepackage[T1]{fontenc}

\usepackage[utf8]{inputenc}

\usepackage{microtype}

\usepackage{inconsolata}

\title{Prompt Tuned Embedding Classification for Industry Sector Allocation}

\author{
  Valentin Leonhard Buchner$^{1,2}$\footnotemark[1] \;
  Lele Cao$^{1}$\thanks{Corresponding authors. The source code is publicly available at \url{https://github.com/EQTPartners/PTEC}.} \;
  Jan-Christoph Kalo$^{2,3}$ \;
  Vilhelm von Ehrenheim$^1$ \\
 $^1$Motherbrain, EQT Group, Stockholm, Sweden \\
 $^2$Vrije Universiteit Amsterdam \; $^3$University of Amsterdam \\
\texttt{\{valentin.buchner,lele.cao,vilhelm.vonehrenheim\}@eqtpartners.com} \;
\texttt{j.c.kalo@uva.nl}
}

\begin{document}
\maketitle
\begin{abstract}
We introduce Prompt Tuned Embedding Classification (PTEC) for classifying companies within an investment firm's proprietary industry taxonomy, supporting their thematic investment strategy. PTEC assigns companies to the sectors they primarily operate in, conceptualizing this process as a multi-label text classification task. Prompt Tuning, usually deployed as a text-to-text (T2T) classification approach, ensures low computational cost while maintaining high task performance. However, T2T classification has limitations on multi-label tasks due to the generation of non-existing labels, permutation invariance of the label sequence, and a lack of confidence scores. PTEC addresses these limitations by utilizing a classification head in place of the Large Language Models (LLMs) language head. PTEC surpasses both baselines and human performance while lowering computational demands. This indicates the continuing need to adapt state-of-the-art methods to domain-specific tasks, even in the era of LLMs with strong generalization abilities.
\end{abstract}

\section{Introduction}

Investors leveraging thematic investment strategies concentrate their efforts on specific industry sectors, such as "Circular Economy." This strategy involves compiling a comprehensive list of companies within these sectors by analyzing unstructured natural language data on platforms such as \citet{Pitchbook2023:2024} and \citet{Crunchbase2023:2024}. For instance, investors might utilize the description and associated keywords of a company like "Vinted" to identify the industries it operates in. In this context, machine learning can be instrumental by framing this as a multi-label text classification challenge: given a natural language description of a company $X$, the goal is to categorize it into one or more industries from a predefined industry sector taxonomy $T = \{Y_1, Y_2, \ldots, Y_n\}$.

While there exist various machine learning solutions for multi-label text classification, this industrial application encompasses some challenges:
\begin{itemize}[leftmargin=*]
\setlength\itemsep{0em}
    \item \textbf{Scarce annotations:} The annotation process, carried out by investment professionals familiar with a firm's taxonomy, results in only a limited number of labeled examples. Given that an industry taxonomy may include up to $300$ industries, there are only few annotations per industry.
    \item \textbf{Imbalanced annotations:} Annotations are primarily focused on investment opportunities relevant to the annotator's industry of interest, leading to a long-tail distribution.
    \item \textbf{Large and heterogeneous inference dataset:} The necessity to infer industries for over $10M$ companies, coupled with the likelihood of the inference data being out-of-distribution compared to the annotated dataset in terms of language use and descriptiveness.
    \item \textbf{Dynamic taxonomy and training data:} Frequent updates in industry taxonomy, company information, and annotations necessitate ongoing re-training and inference processes.
\end{itemize}

Traditional text classification approaches demand large amounts of annotated training data and often struggle to generalize effectively to novel data~\citep{srivastava2022beyond:2022}. Large Language Models (LLMs) exhibit superior generalization capabilities to unseen data and can be fine-tuned on smaller annotated datasets~\citep{raffel2020exploring:2020}. However, fine-tuning LLMs may lead to the undesirable phenomenon of "catastrophic forgetting" of pretraining knowledge~\citep{chen2020recall:2020}, and is computationally demanding. These challenges can be mitigated through Parameter-Efficient Fine-Tuning (PEFT,~\citealp{ding2023parameter:2023}) techniques such as Prompt Tuning (PT). PT minimizes the number of parameters that need fine-tuning by focusing on a \textit{soft prompt} appended to the tokenized and embedded input text, thus reducing computational costs and preserving the pretrained knowledge of the LLM, as the main body of the LLM's parameters remains unaltered~\citep{tam2022parameter:2022, tu2022prompt:2022, lester2021power:2021}. Hence, PT emerges as a viable solution for computational efficiency and knowledge retention in LLM applications.

This research evaluates the scalability, efficiency, and performance of PT in a real-world industry classification scenario, benchmarked against common baseline methods. However, PT as a text-to-text (T2T) classification approach encounters limitations on multi-label tasks as discussed in Subsection \ref{sec:prompt_tuning}. We enhance PT by (1) integrating constrained decoding using Trie Search~\citep{yang2022multi:2022, de2020autoregressive:2020} and (2) replacing the language model head with a specialized classification head. Our key contributions include:
\begin{itemize}[leftmargin=*]
\setlength\itemsep{0em}
    \item The adaptation of the Trie Search decoding method~\citep{yang2022multi:2022}, preventing repetitive prediction of the same label, akin to the approach in~\citep{chen2018order:2018}.
    \item The introduction of Prompt Tuned Embedding Classification (PTEC), which concurrently optimizes the \textit{soft prompt} and the classification head with differential learning rates.
    \item A comparative analysis of the performance and computational requirements of the proposed and baseline methods on two datasets: our proprietary \textit{IndustrySector} classification task and the publicly available \textit{HateSpeech} classification benchmark.
    \item Empirical evidence demonstrating that evaluating PTEC on data it has more pretraining knowledge about does not lead to an overestimation of the its classification performance when applied to data it has less pretraining knowledge about.
\end{itemize}

The paper first outlines existing text classification methodologies and their limitations. We then introduce constrained Trie Search decoding and PTEC as potential solutions to these limitations. Subsequently, we describe our experimental setup and compare the efficiency and performance of current and proposed methods. Our codebase and the \textit{HateSpeech} dataset can be accessed at \url{https://github.com/EQTPartners/PTEC}.

\section{Related Methods}

\subsection{Parameter-Free Classification with gzip}

A very simple approach to text classification makes use of compression algorithms such as gzip~\cite{jiang-etal-2023-low:2023}. This method leverages the principle of lossless compression, where frequently occurring symbols are encoded with shorter codes. Similar texts are likely to have more common symbols, leading to a shorter compressed length when concatenated. This phenomenon forms the basis for a low-computation distance metric for nearest-neighbors classification methods.

\subsection{In-Context Learning}
\label{icl}

In-Context Learning (ICL), or $N$-shot prompting, involves prepending $N$ input-output example pairs to the prompt before the actual input~\cite{brown2020language:2020, min2022rethinking:2022}. This method is particularly appealing for text classification as it obviates the need for any LLM fine-tuning.

\subsection{Embedding Proximity}

Another approach to text classification not requiring LLM fine-tuning uses text embeddings generated with LLMs. These can be used as input features for a separate classification model. The most parameter-efficient classification models are K-Nearest Neighbors (KNN) or Radius Nearest Neighbors (RadiusNN)~\cite{guo2003knn:2003, nnc:1967}. Alternatively, text embeddings can be used as input to a classification layer, which can be trained to perform the respective classification task~\cite{kowsari2019text:2019}.

\subsection{Prompt Tuning}
\label{sec:prompt_tuning}

\begin{figure}
    \centering
    \includegraphics[width=0.49\textwidth]{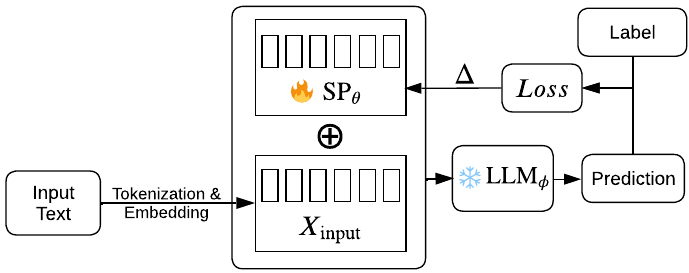}
    \caption{Schematic overview of Prompt Tuning, showing the trainable \textit{soft prompt}
(matrix $SP_\theta$), the tokenized and embedded input text ($X_\text{input}$), and the LLM with frozen parameters ($LLM_\phi$).}
\label{fig:prompt_tuning}
\end{figure}

To emulate fine-tuning's effectiveness with reduced computational expense, various Parameter-Efficient Fine-Tuning (PEFT) techniques have been developed. These include Pattern-Exploiting Training~\cite{schick2020s:2020}, Prefix-Tuning~\cite{li2021prefix:2021}, Low-Rank Adaptation (LoRa,~\citealp{hu2021lora:2021}), and Prompt Tuning~\cite{lester2021power:2021, liu2021p:2021, tam2022parameter:2022}. These methods limit trainable parameters compared to full LLM fine-tuning. PT involves training the smallest amount of parameters ($<0.1\%$), while still being reported to outperform fine-tuning~\cite{liu2021gpt:2021}. It prepends a \textit{soft prompt} — a sequence of virtual token embeddings — to the token embeddings of the input text, as depicted in Fig. \ref{fig:prompt_tuning}. During this process, only the \textit{soft prompt} undergoes training, leaving the LLM's parameters unchanged. This approach not only demands fewer computational resources but also supports multi-task processing in a single batch and mitigates the risk of "catastrophic forgetting."

\subsection{T2T Classification for Multi-Label Tasks}
\label{sec:multilabelt2t}

Text-to-Text (T2T) classification leverages generative language models to produce the token(s) representing target categories. Historically, T2T has surpassed other methods in public benchmarks, aligning with the notion that text generation closely mirrors the LLM's pretraining tasks~\cite{raffel2020exploring:2020}. For multi-label scenarios, T2T classification sequentially generates labels, separated by a separator token (SEP) and concluded with an end-of-sequence (EOS) token~\cite{yang2018sgm:2018, yang2022multi:2022}. However, this approach faces several limitations:
(a)\label{limitation_incorrect} The model might generate semantically similar but incorrect labels due to non-intuitive class labels. For instance, in our proprietary taxonomy, the model could misclassify "Healthcare IT" as "Healthcare Software". (b)\label{limitation_order} In multi-label instances, labels must be provided in an arbitrary order during fine-tuning. If the model's correct label predictions deviate from this order, it is penalized by the loss function. Augmenting the label order at random would result in an inconsistent learning signal and unstable convergence. (c)\label{limitation_confidence} The model computes the probability of a subsequent label based on the previously decoded label, expressed as $P(Y_2|X, Y_1)$, where $X$ is the input and $Y_i$ represents the $i$-th label~\cite{simig2022open:2022}. This approach fails to provide independent confidence scores for each label $P(Y_2|X)$, which are vital in real-world applications for balancing the trade-off between false positives and false negatives. Additionally, this limitation does not allow for achieving optimal performance in metrics like Precision@K, which depend on label probabilities.

\section{Proposed Methods}
\label{sec:prop}

\subsection{Prompt Tuning + Trie Search}

\begin{figure*}
    \centering
    \includegraphics[width=0.99\textwidth]{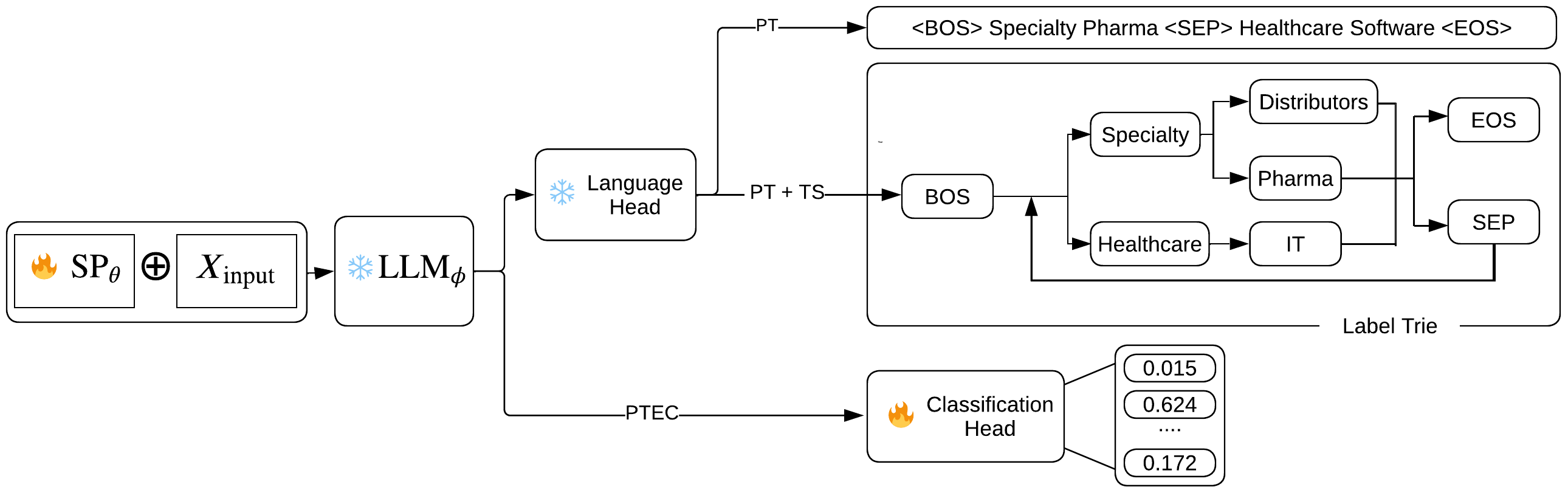}
    \caption{A schematic comparison of Prompt Tuning with T2T classification (PT + T2T), Prompt Tuning with Trie Search (PT + TS), and PTEC. Note that \textit{Healthcare Software} would not be a valid label name, while \textit{Healthcare IT} would be.}
    \label{fig:comparison}
\end{figure*}

To address limitation (a) as detailed in Section \ref{sec:multilabelt2t}, constrained decoding methods such as Trie Search, which are effective in generating only valid labels, can be employed \cite{de2020autoregressive:2020, yang2022multi:2022}. Trie Search, a constrained decoding method, utilizes a label trie structure for organizing target labels, as illustrated in Fig. \ref{fig:comparison}. The label trie, beginning from the root node (BOS) and ending at leaf nodes (EOS or SEP), enables valid label retrieval during label generation by guiding the LLM to select tokens only from the trie. In the context of multi-label classification, labels are generated sequentially and separated by the SEP token. Upon reaching a leaf node, the LLM chooses either to generate the SEP token, restarting the Trie Search, or the EOS token, concluding label prediction. However, this method may lead to repetitive generation of the same label, a known issue with LLMs \cite{fu2021theoretical:2021}. To mitigate this, our approach extends the Trie Search method by removing a label from the trie once it is generated, an idea inspired by \cite{chen2018order:2018}. While this method effectively addresses limitations (a), it does not resolve limitations (b) and (c) since it requires labels provided in an arbitrary order during training and does not allow the calculation of appropriate confidence scores.

\subsection{Prompt Tuned Embedding Classification}
PTEC addresses all limitations by combining PT with Embedding Classification rather than T2T classification. This is done by using a single linear layer with a sigmoid activation function to process the text embeddings generated by the Prompt Tuned LLM. This layer produces a probability distribution over industry sectors in the taxonomy, thus (a) ensuring valid industry selection, (b) enabling the application of label order-independent loss functions, and (c) providing probability scores useful for ranking or adjusting model prediction sensitivity. This process is mathematically represented as:
\begin{equation}
    \label{eq:ptec}
    p = 
\begin{cases} 
1 & \text{if } \sigma(\mathbf{W} \text{LLM}_\phi(\text{SP}_\theta \oplus \mathbf{X}_\text{input}) + \mathbf{b}) \geq \tau, \\
0 & \text{otherwise.}
\end{cases} 
\end{equation}
Here, $\text{LLM}_\phi(\text{SP}_\theta \oplus \mathbf{X}_\text{input})$ parameterized by $\phi$ yields an embedding vector. The tokenized and embedded input text is represented by $\mathbf{X}_\text{input}$, and $\tau$ is the
threshold used. The weight matrix $\mathbf{W} \in R^{d \times l}$ and bias vector $\mathbf{b} \in R^{l}$ are components of the linear layer, with $d$ representing the dimensionality of the LLM's embedding vector and $l$ the number of labels. During training, the task-specific classification layer and the \textit{soft prompt} are optimized concurrently, while the rest of the LLM's parameters are kept frozen. This approach is akin to strategies used in Named Entity Recognition \cite{liu2021p:2021} and multi-class text classification \cite{hambardzumyan2021warp:2021}. Following the observation by \citet{lester2021power:2021}, we found that a \textit{soft prompt} typically benefits from a higher learning rate, while the classification head performs optimally with a lower rate. Hence, in our PTEC implementation, differential learning rates are applied to the \textit{soft prompt} and the classification head. Besides addressing the limitations listed in Section \ref{sec:prompt_tuning}, PTEC offers the advantage of faster inference times, requiring only a single forward pass per prediction compared to one forward pass for each predicted token.

\section{Experiments}

\subsection{Dataset}
\label{sec:datasets}

Based on an investment firm's proprietary database we constructed the \textit{IndustrySector} dataset of around $5500$ companies. Each company is annotated with $1$ to $4$ of $76$ different industries, and each industry is labeled at least $20$ times. For each company, its legal name, keywords, and a description are available. This information is concatenated to one text used as the input prompt in all experiments. Appendix \ref{app:indsec} describes dataset analytics and preprocessing steps. To facility reproducibility, we further constructed the public \textit{HateSpeech} benchmark, which is elaborated on in Appendix~\ref{app:benchmark}.

\subsection{Model Training}

Our PT set-up follows the architecture described in Section \ref{sec:prop}. Since for T2T classification the labels need to be provided in a predefined order during training, we sort the labels for each sample descending by their frequency in the training data as this has been confirmed to provide the best performance~\cite{yang2018sgm:2018, jung2023cluster:2023}. We noticed that classes with class labels consisting of more tokens have more influence on the cross-entropy loss than classes with shorter labels. Consequently, we developed the Normalized Token Entropy (NTE) Loss, which is motivated and elaborated on in Appendix \ref{app:nte}. Further, we use token embeddings of the target classes to initialize the \textit{soft prompt}'s weights, as~\citet{lester2021power:2021} showed this to be beneficial for task performance. As there are more tokens available for the target classes than there are tokens in the \textit{soft prompt}, we randomly sample the tokens to be used for \textit{soft prompt} initialization. All methods are compared using the $7B$ parameter version of LLaMa (LLaMa 7B,~\citealp{touvron2023llama:2023}) and the $1.7B$ parameter version of Bloom (Bloom 1B7,~\citealp{scao2022bloom:2022}). A detailed description of our hyperparameter tuning strategy can be found in Appendix \ref{app:hyptun}.

\subsection{Metrics}
\label{sec:metrics}

To achieve optimal business impact, it is crucial to predict all industry sectors similarly well. This enables an investment firm to not only find companies in well-explored sectors but also in novel or niche sectors. Consequently, we use the macro-averaged F1 score to compare model performance. Further, it becomes important to be cost-effective when frequently retraining and running inference over a large database. Therefore, we report on the computational resources required for fine-tuning and for inference over $10M$ companies by measuring the consumed floating point operations (FLOPs). These were measured using Pytorch's profiler~\cite{PyTorch2023:2024} for a representative sample of batches, and the results were extrapolated on the full training and inference process. The FLOPs consumption of KNN and RadiusNN were estimated as motivated in Appendix~\ref{app:knnflops}. To investigate the subjectivity of this industry classification task, an exhaustive list of labels was created for a representative subsample of the test set ($N = 104$) and annotated by $3$ independent professional raters. Chance-corrected inter-annotater agreement was calculated using Cohen's kappa ($\kappa$,~\citealp{mchugh2012interrater:2012}).

\subsection{The Impact of Pretraining Knowledge}

Companies in our \textit{IndustrySector} dataset were annotated depending on investment professionals' interests and are not a representative sample of the inference dataset. On the contrary, investment professionals are more likely to annotate companies that are more widely known, which are companies the LLM may have encountered during pretraining. The LLM may thus perform the desired downstream task better for the annotated companies in our test set than for the full set of less-known companies in the inference dataset, resulting in an overestimation of model performance. To investigate whether this is the case, we prompted an LLM to indicate about which companies in the test set it has pretraining knowledge, following the logic that LLMs mostly know what they know~\cite{kadavath2022language:2022}. We then conducted a nonparametric Mann-Whitney U test~\cite{nachar2008mann:2008} to test the hypothesis $H_1$ that \textit{classification performance is higher for the companies the LLM indicates to have pretraining knowledge about}.

\section{Results}
\label{sec:results}

\subsection{Performance and Computational Cost}

The computational efficiency and average performance over $3$ runs of various methods on the \textit{IndustrySector} dataset are presented in Table \ref{table:perf_flops}. PTEC shows an improvement ranging from 3.6 to 11.7 percentage points over the next best method while being more efficient than other PT methods for both training and inference. Additionally, PTEC shows less variability between runs than PT with T2T classification, particularly for Bloom 1B7.

\begin{table}
\centering
\small
\setlength{\tabcolsep}{4pt}
\begin{tabular}{llrrrr}
\toprule
& \textbf{Method} & \multicolumn{2}{c}{\textbf{FLOPs}} & \multicolumn{2}{c}{\textbf{Macro F1}} \\
\cmidrule(lr){3-4} \cmidrule(lr){5-6}
& & \multicolumn{1}{c}{\textbf{Training}} & \multicolumn{1}{c}{\textbf{Inference}} & \multicolumn{1}{c}{\textbf{Mean}} & \multicolumn{1}{c}{\textbf{Std}} \\
\midrule
\multirow{8}{*}{\rotatebox{90}{Bloom 1B7}} & PTEC & 1.12e+17 & 1.09e+18 & \textbf{0.398} & 0.019 \\
& PT + TS & 8.96e+16 & 1.65e+18 & 0.240 & 0.060 \\
& PT & 8.96e+16 & 1.65e+18 & 0.221 & 0.068 \\
& CH & 3.29e+16 & \textbf{3.97e+17} & 0.281 & 0.006 \\
& KNN & 3.29e+16 & \textbf{3.97e+17} & 0.230 & 0 \\
& RadiusNN & 3.29e+16 & \textbf{3.97e+17} & 0.101 & 0 \\
& $N$-shot + TS & \textbf{0} & 8.51e+18 & 0.134 & 0.004 \\
& $N$-shot & \textbf{0} & 5.68e+18 & 0.025 & 0.005 \\
\addlinespace
\multirow{8}{*}{\rotatebox{90}{LLaMa 7B}} & PTEC & 1.69e+17 & 4.27e+18 & \textbf{0.448} & 0.001 \\
& PT + TS & 9.73e+17 & 5.62e+18 & 0.412 & 0.005 \\
& PT & 9.73e+17 & 5.62e+18 & 0.412 & 0.002 \\
& CH & 2.13e+17 & \textbf{2.56e+18} & 0.400 & 0.007 \\
& KNN & 2.13e+17 & \textbf{2.56e+18} & 0.332 & 0 \\
& RadiusNN & 2.13e+17 & \textbf{2.56e+18} & 0.237 & 0 \\
& $N$-shot + TS & \textbf{0} & 2.59e+19 & 0.032 & 0.001 \\
& $N$-shot & \textbf{0} & 2.55e+19 & 0.015 & 0.002 \\
\addlinespace
- & gzip & $-$ & $-$ & 0.271 & 0 \\
\bottomrule
\multicolumn{6}{l}{\footnotesize\parbox{0.95\linewidth}{CH = classification head; gzip = parameter-free classification with gzip. Other abbreviations as defined in Fig. \ref{fig:comparison}.}}
\end{tabular}
\caption{Results on the \textit{IndustrySector} dataset. The method with the lowest FLOPs and highest Macro F1 Score is highlighted in \textbf{bold} for each LLM. A dash ($-$) indicates unavailable data or no LLM required.}
\label{table:perf_flops}
\end{table}

Contrasting prior findings where T2T classification outperformed classification heads~\cite{raffel2020exploring:2020}, PTEC outperforms PT + T2T in our study. Several arguments can be made to explain this: (1) T2T classification often outperforms because the LLM can make a reasonable guess. However, the proprietary and domain-specific nature of the industry taxonomy limits the LLM's ability to leverage its pretraining knowledge. (2) While most tasks used to evaluate T2T classification can be reduced to singular-token targets ("good" or "bad"), the \textit{IndustrySector} dataset consists of multi-token labels and therefore presents a more complex label space.

Trie Search enhances T2T classification performance by 0.17 to 10.9 percentage points with $N$-shot prompting. However, it does not improve LLaMa 7B's performance when used with PT, suggesting that PT effectively learns to predict valid labels such that Trie Search does not result in any additional performance gain.

Classification heads demonstrate comparable performance to PT with T2T classification but are significantly more computationally efficient. While $N$-shot prompting eliminates training FLOPs, it necessitates a higher number of inference FLOPs. Table \ref{table:ablation} summarizes the techniques each method employs. Results on our public \textit{HateSpeech} benchmarking dataset followed nearly the same pattern and can be inspected in Appendix \ref{app:benchmark}.

\begin{figure}
  \centering
  \includegraphics[width=0.45\textwidth]{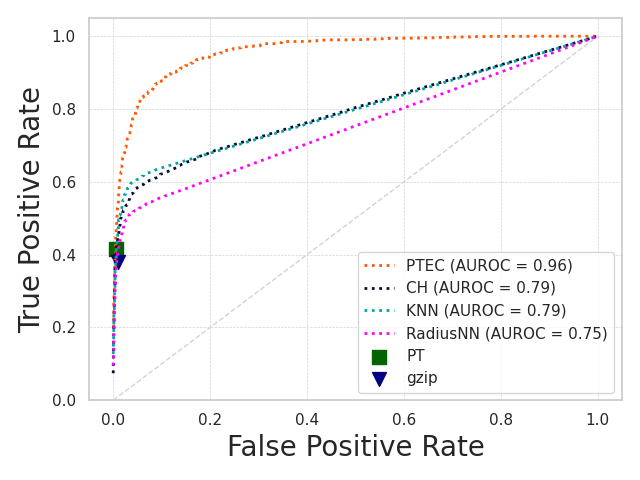}
  \caption{ROC curves using LLaMa 7B. Methods that cannot be thresholded are displayed as individual points. AUROC = Area Under the ROC curve. Other abbreviations as defined in Fig. \ref{fig:comparison} and Table \ref{table:perf_flops}.}
  \label{fig:roc}
\end{figure}

Methods such as PTEC offer the advantage of predicting appropriate confidence scores. This attribute is evident in Fig. \ref{fig:roc}, which displays the Receiver Operating Characteristic (ROC) curves for multiple methods. These confidence scores allow for selecting a threshold to choose the appropriate trade-off between precision and recall, a crucial attribute for deploying a model in production.

\begin{table}
\centering
\small
\setlength{\tabcolsep}{4pt}
\begin{tabular}{lcccccccc}
\toprule
& valid & order & conf. & LLM & Macro \\
& labels & invariant & scores & tuning & F1\\
\cmidrule(lr){2-5}\cmidrule(lr){6-7}
$N$-shot & & \checkmark & & & 0.015\\
$N$-shot + TS & \checkmark & \checkmark & & & 0.032\\
RadiusNN & \checkmark & \checkmark & \checkmark & & 0.237\\
KNN & \checkmark & \checkmark & \checkmark & & 0.332\\
CH & \checkmark & \checkmark & \checkmark & & 0.4\\
PT + T2T & & & & \checkmark & 0.412\\
PT + TS & \checkmark & & & \checkmark & 0.412\\
PTEC & \checkmark & \checkmark & \checkmark & \checkmark & \textbf{0.448}\\
\bottomrule
\multicolumn{9}{l}{\footnotesize\parbox{0.95\linewidth}{Abbreviations as defined in Fig. \ref{fig:comparison} and Table \ref{table:perf_flops}}}
\end{tabular}
\caption{Overview of methods used and their performance on the \textit{IndustrySector} dataset using LLaMa 7B. The highest F1 score is highlighted in \textbf{bold}.}
\label{table:ablation}
\end{table}

\subsection{The Impact of Pretraining Knowledge}

In the \textit{IndustrySector} dataset's test split, $159$ of the $839$ companies were recognized from pretraining, while $680$ were not. A qualitative review confirmed that known companies had more accessible online information than unknown companies. A Mann-Whitney U test indicated that differences in task performance using LLaMa 7B between both groups were nonsignificant at a p-value of $0.243$ (U = $50993.5$; r = $0.0385$). This results in the rejection of $H_1$ that \textit{classification performance is higher for the companies the LLM indicates to have pretraining knowledge about}. This indicates that we likely do not overestimate performance on the inference dataset.

\subsection{Inter-rater Agreement}

Table \ref{tabel:agreement} displays the interrater agreement between three independent human raters, the gold labels used to train PTEC, and PTEC predictions on the subsample described in Section \ref{sec:metrics}. The moderate agreement between human raters verifies the subjectivity of our \textit{IndustrySector} classification task. Out of $104$ companies, unanimous agreement was reached on just $6$ companies. Importantly, PTEC's agreement with the gold labels is up to 15.1 percentage points higher than the agreement between human raters and the gold labels. This shows that PTEC outperforms human professionals, meaning that it provides value by accelerating and objectifying the industry classification process.

\begin{table}
\centering
\small
\setlength{\tabcolsep}{4pt}
\begin{tabular}{lccccc}
\toprule
         & Rater2   & Rater3   & Gold   &   PTEC & $\Delta_{	\text{Gold} - \text{PTEC}}$$^{\mathrm{a}}$       \\
\midrule
 Rater1  & 0.477    & 0.401    & 0.389  &  0.36  & 0.029 \\
 Rater2  &          & 0.444    & 0.551  &  0.422 & 0.129 \\
 Rater3  &          &          & 0.311  &  0.245 & 0.066 \\
 Average &          &          & 0.417  &  0.342 & 0.075 \\
 \midrule
 Gold    &          &          &        &  0.562 &       \\
\bottomrule
\multicolumn{6}{l}{\footnotesize\parbox{1\linewidth}{$^{\mathrm{a}}$the difference in agreement of a given rater with the gold annotations and the PTEC predictions.}}
\end{tabular}
\caption{Agreement Matrix using Cohen's Kappa comparing three independent human raters, gold labels, and predictions made with PTEC LLaMA 7B.}
\label{tabel:agreement}
\end{table}

\section{Conclusion}


This study benchmarks computational cost and multi-label text classification performance of PT as a parameter-efficient alternative to fine-tuning all LLM parameters. To address the limitations of a T2T approach on multi-label classification problems, PT is extended with Trie Search as a constrained decoding strategy, and with Embedding Classification as an alternative to T2T classification. 
Results indicate that Trie Search can significantly improve the performance of $N$-shot prompting. PT can outperform popular text classification approaches on both our domain-specific \textit{IndustrySector} classification task, and the publicly released \textit{HateSpeech} classification benchmark. Both performance and efficiency can be further improved by combining PT with Embedding Classification. The proposed solution, PTEC, outperforms baselines and human professionals and can be deployed at scale to accelerate and objectify industry sector allocation.

\bibliography{ptec}

\newpage

\appendix

\section{Appendix}
\label{sec:appendix}

\subsection{Inference FLOPs Calculation for Nearest-Neighbors Methods}
\label{app:knnflops}

KNN and RadiusNN were implemented using sklearn \citep{scikit-learn:2011}. There is to our knowledge no existing method to measure their FLOPs consumption for nearest-neighbor methods implemented with sklearn during inference. Instead, their inference FLOPs were estimated as:
\begin{equation}
\label{eq:flops}
\text{FLOPs} \approx E(T + I) + 3(D\cdot T\cdot I)
\end{equation}

Here, $D$ represents the dimensionality of the text embeddings, $T$ denotes the number of training samples, $I$ indicates the number of inference samples, and $E$ the FLOPs required to embed one example. This equation can be derived as follows: The term $E(T + I)$ refers to calculating the embeddings for the training and inference samples, and $3(D \cdot T \cdot I)$ estimates the number of floating point operations (FLOPs) for performing classification with the KNN and RadiusNN algorithms. The average value of $E$ is calculated by measuring the FLOPs used for generating one embedding with PyTorch's profiler.
Assuming a brute-force implementation, for both KNN and RadiusNN, each inference embedding is compared with every training embedding.
The term $3 \cdot D$ corresponds to calculating the Euclidean distance between two embeddings. This calculation involves the subtraction of one embedding from the other ($D$ FLOPs), squaring each element of the new vector ($D$ FLOPs), taking the sum of these values ($D-1$ FLOPs) and finally taking the square root of this sum ($1$ FLOP). As this is done once for each pair of training and inference examples, the distance calculations will need $3(D \cdot N \cdot M)$ FLOPs in total. 


As this is only an estimate, the exact number can vary based on the specifics of the operations used. While the formula provided here assumes a brute-force method for KNN and RadiusNN, it is important to note that more efficient methods are often employed in practice, especially in popular machine learning libraries such as scikit-learn~\citep{scikit-learn:2011}. True computational resources required by KNN and RadiusNN methods may therefore be lower than estimated in this paper. However, this estimation provides a general idea of the computational resources needed. For both RadiusNN and KNN the FLOPs used for calculating the text embeddings of the training data are considered as `training' FLOPs. 

\subsection{IndustrySector Dataset Preprocessing}
\label{app:indsec}

\begin{figure*}
  \centering
  \subfloat{
    \includegraphics[width=0.23\textwidth]{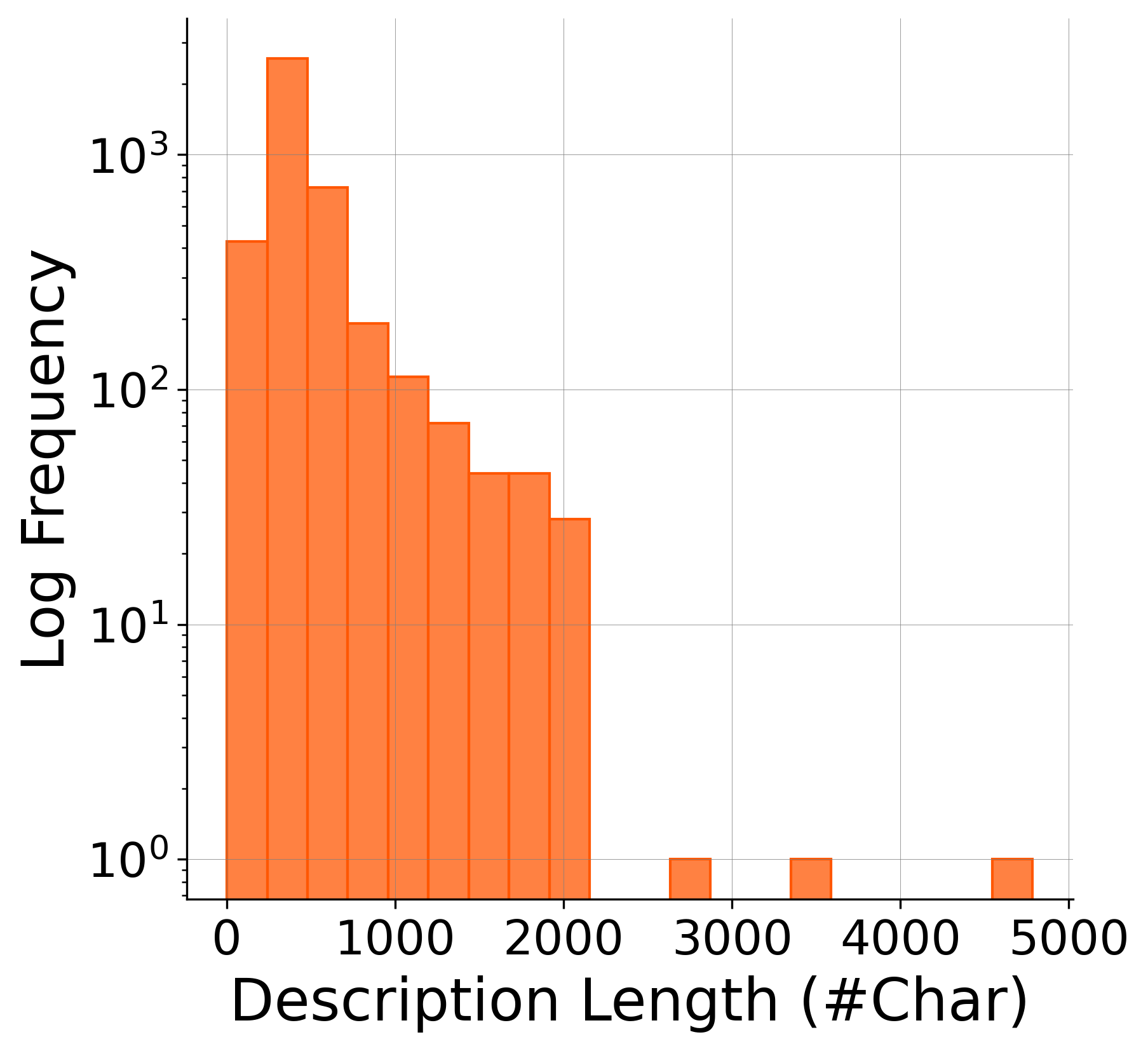}%
    \label{fig:des_len_og}
  }
  \subfloat{
    \includegraphics[width=0.23\textwidth]{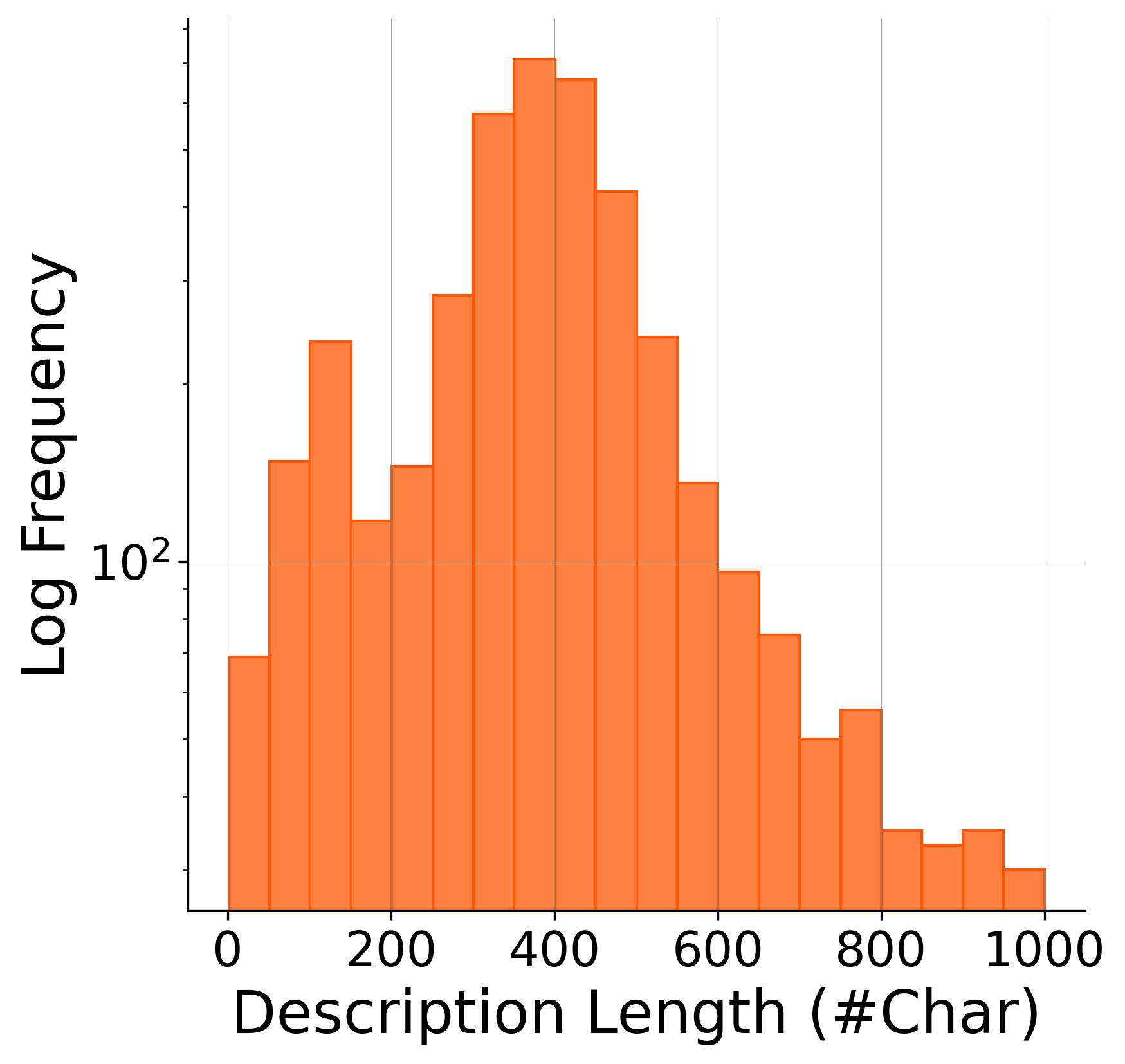}
    \label{fig:des_len_pre}
  }
  \subfloat{
    \includegraphics[width=0.23\textwidth]{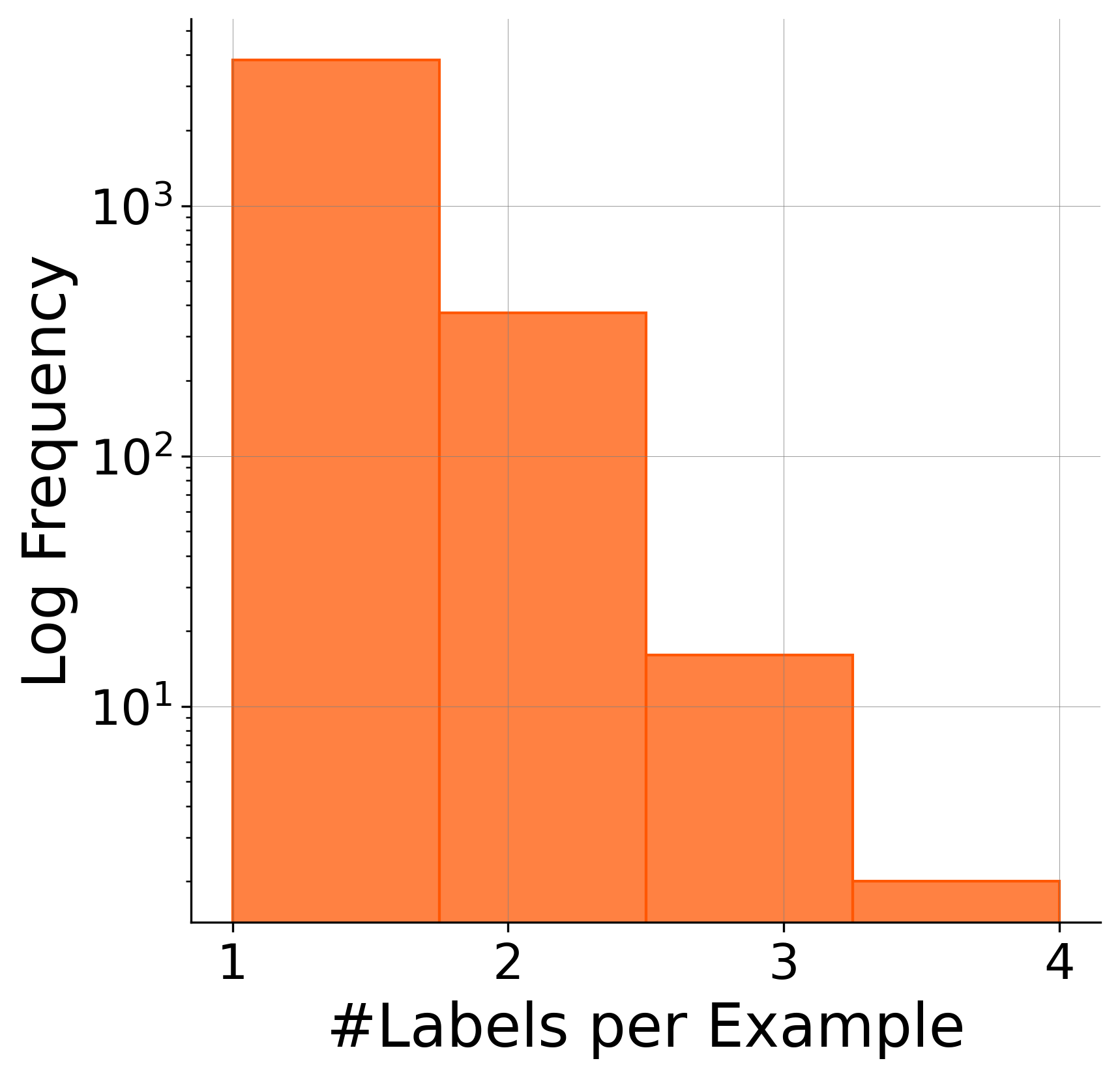}
    \label{fig:label_count}
  }
  \subfloat{
    \includegraphics[width=0.23\textwidth]{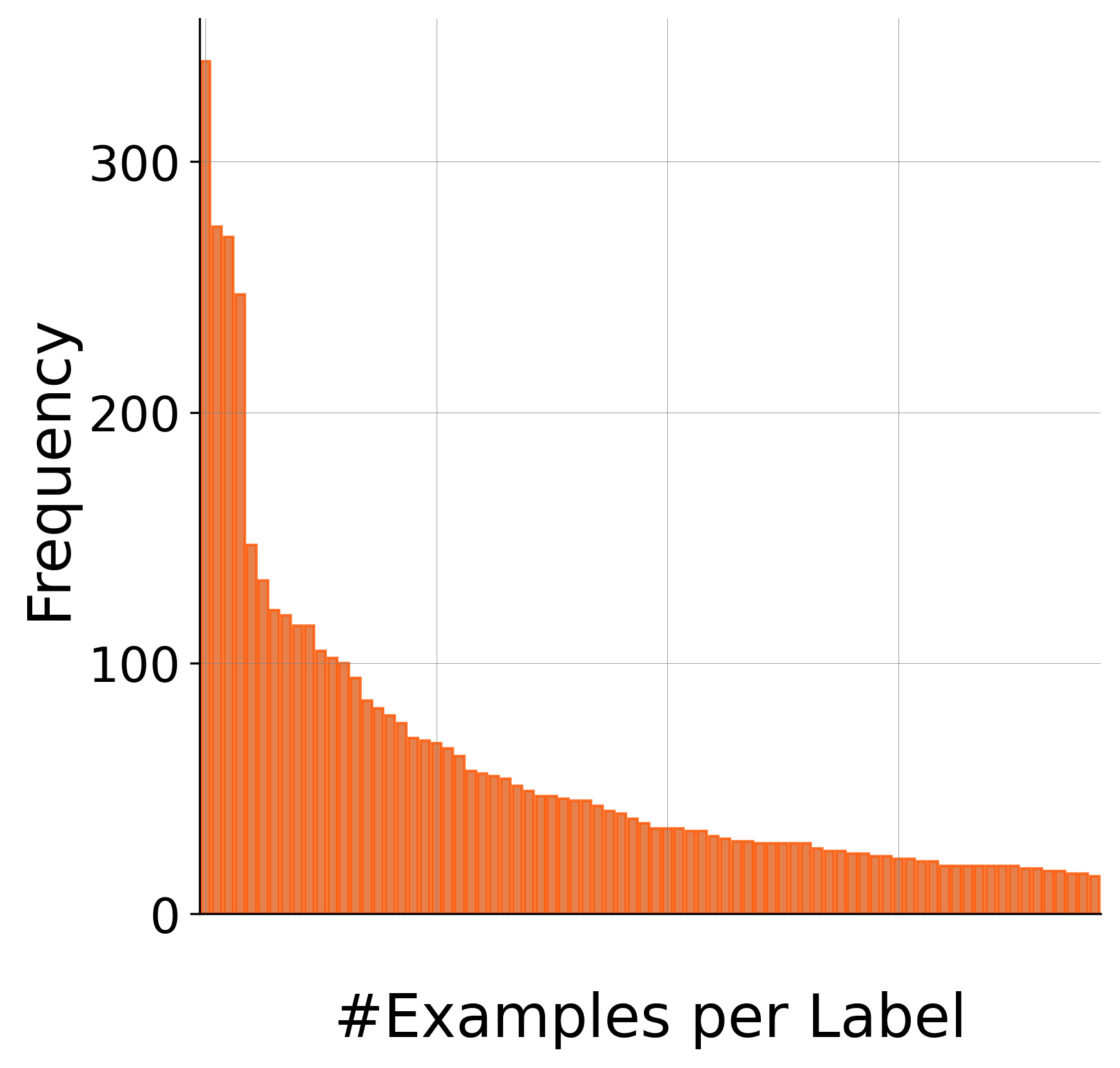}
    \label{fig:label_dist}
  }
  \caption{Distributions of (a) original description lengths, (b) preprocessed description lengths, (c) number of labels per example, and (d) number of examples per label}
  \label{fig:dis}
\end{figure*}

The average number of labels in the \textit{IndustrySector} dataset per example is $1.1$. This indicates that while the problem, in theory, is a multi-label classification problem, most examples in our dataset are not exhaustively annotated and only carry one label (see Fig. \ref{fig:dis}).
The dataset is split into $75\%$ training set, $10\%$ validation set, and $15\%$ test set. Fig. \ref{fig:dis} shows the highly imbalanced, long-tail class distribution: some industries occur only $\sim 25$ times, while the most frequent industry occurs $>300$ times. Importantly, this distribution only shows the classes included in the \textit{IndustrySector} dataset, and our database contains many more classes with even fewer annotations. 
To ensure that each industry in the \textit{IndustrySector} dataset is represented in similar proportions in all splits, and with a minimum frequency in both validation and test split, stratification is performed using multi-label stratified shuffle splitting, as proposed by~\citet{sechidis2011stratification:2011}. During this process, it is ensured that each industry is represented at least $2$ times in the validation set, $3$ times in the test set, and $15$ times in the training set. The imbalanced annotations were accounted for by reweighing the loss: Class weights are calculated for each class with 
$n_\text{max}/n_\text{class}$. The loss for each instance is weighted by its class weight before updating the gradients.

Since the LLM's self-attention mechanism's complexity increases quadratically with prompt length, long input prompts will easily result in out-of-memory (OOM) errors. Therefore, descriptions and keyword lists that consist of more than $1000$ characters are summarized using the $250M$ parameter instruction fine-tuned FLAN T5 model~\citep{chung2022scaling:2022}, such that no input prompt supersedes a length of $1000$ characters. The result of this summarizing step is displayed in Fig. \ref{fig:dis}.

\subsection{Normalized Token Entropy (NTE) Loss}
\label{app:nte}

Careful attention has to be paid to the loss calculation when performing mini-batch gradient descent. As PyTorch's~\cite{PyTorch2023:2024} cross-entropy loss function by default averages the loss over all label tokens in a batch, industries with names consisting of more tokens ("Circular Economy \& Sustainable Materials") have a larger influence on the batch loss than industries with shorter names ("Marketplaces"). 
This results in the model learning industries with longer names better than industries with shorter names. To avoid this, we adjust the cross-entropy loss calculation such that each label has the same influence on the batch loss by reweighting the influence that each token has on the loss. This can be done by first taking the average loss of all tokens belonging to one label, and then averaging all individual losses over the batch. This is denoted in \eqref{eq:loss}, where $L$ is the aggregated loss of the batch, $N$ is the number of examples of the batch, $y_{i}$ is the label tokens for the i-th example in the batch, $|y_{i}|$ is the length of the label of the i-th example measured in it's number of tokens, $y_{ij}$ is the target value of the j-th token of the i-th label, and $p_{ij}$ is the predicted probability of the j-th token of the i-th label.
\begin{equation}
\label{eq:loss}
    L = -\frac{1}{N}\sum_{i=1}^{N} \left(\frac{1}{|y_{i}|}\sum_{j=1}^{|y_{i}|} y_{ij} \log(p_{ij})\right)
\end{equation}

\vspace{1mm}

\subsection{Hyperparameter Tuning}
\label{app:hyptun}

The hyperparameters for all methods were optimized using Bayesian Optimization~\cite{snoek2012practical:2012} with $25$ random initializations of hyperparameter combinations and $15$ iterations of Bayesian Optimization. Models involving PT are trained using the AdamW optimizer. Hyperparamters such as the learning rate and weight decay were searched on a logarithmic scale, such that the probability to sample values from the interval $[ 0.01 \leq x \leq 0.1 ]$ equals the probability to sample values from $[ 0.001 \leq x \leq 0.01 ]$, given that both intervals are included in the searched hyperparameter space. For the KNN and RadiusNN methods, the optimal hyperparameter values have large variability between different models. For this reason, if a hyperparameter was close to the boundary of the searched hyperparameter space, Bayesian Optimization was continued with a broader hyperparameter range. An overview over the optimized hyperparamters, the scale of searching, and the ranges of hyperparameter values searched are provided in Table \ref{table:hyperparam}. Hyperparameter tuning was performed using the validation set, while all results reported in Section \ref{sec:results} were calculated over the test set. While the maximum batch size fitting on one A100 GPU was used for model training, an effective batch size of $32$ was used for gradient updates. Threshold $\tau$ mentioned in \eqref{eq:ptec} is not considered a hyperparameter, since we automatically select the value that optimized the F1 score. 

\begin{table}
\centering
\setlength{\tabcolsep}{4pt}
\small
\begin{tabular}{lllcc}
\toprule
\textbf{Method}           & \textbf{Hyperp}& \textbf{Scl} & \textbf{Searched Space}&\textbf{Value}\\ \midrule
$N$-shot                               & n                                                   & lin                    & $\{0, 1, ..., 8\}$                   &$7$\\ 
\addlinespace
RadiusNN                             & radius                                              & lin                    & $[0.1, 150]$                          &$25.25$\\ 
\addlinespace
KNN                                  & k                                                   & lin                   & $\{1, 2, ..., 150\}$                  &$1$\\ 
\addlinespace
\multirow{2}{*}{CH} & lr                                       & log                       & $[1e^{-6}, 1e^{-3}]$                  &$1e^{-3.58}$\\ 
& wd                                        & log                       & $[0, 1e^{-3}]$                        &$0$\\ 
\addlinespace
\multirow{3}{*}{PT (+ TS)}                        & SP lr & log                       & $[1e^{-9}, 1]$                      &$1e^{-1.66}$\\ 
& SP length        & lin                    & $\{50, 51, ..., 200\}$                &$156$\\ 
& epochs                                              & lin                    & $\{5, 6, ..., 18\}$                   &$18$\\ 
\addlinespace
\multirow{5}{*}{PTEC}                                 & SP lr & log                       & $[1e^{-9}, 1]$&$1e^{-4.95}$\\ 
& SP length        & lin                    & $\{50, 51, ..., 200\}$&$53$\\ 
& CH lr                                    & log                       & $[1e^{-9}, 0.1]$&$1e^{-4.23}$\\ 
& wd                                        & log                       & $[1e^{-9}, 0.5]$&$1e^{-8.72}$\\
& epochs                                              & lin                    & $\{5, 6, ..., 18\}$ &$13$ \\
\bottomrule
\multicolumn{5}{l}{\footnotesize\parbox{0.95\linewidth}{Abbreviations as defined in Fig. \ref{fig:comparison} and Table \ref{table:perf_flops}}}
\end{tabular}
\caption{Overview of hyperparameters (hyperp), scales (scl), and search space. To ensure reproducibility, value refers to the selected value for LLaMa 7B on the public HateSpeech dataset.}
\label{table:hyperparam}
\end{table}

\subsection{Public Benchmarking}
\label{app:benchmark}

To enable reproducibility, we constructed a public benchmark from \citeposs{salminen2018anatomy:2018} hatespeech classification dataset. The task of this dataset is to classify social media comments into different kinds of hatespeech, where each comment can have one or multiple labels. This dataset was chosen because it is structurally similar to our \textit{IndustrySector} dataset: It covers a set of $22$ different classes, its data is highly imbalanced, and the length of the social media comments is similarly distributed as the length of the company descriptions. Each hate speech comment is annotated with $1$ to $4$ labels, and a comment has $1.45$ annotations on average. It should be noted that we could only find a substantially smaller and differently distributed subset of the original dataset, implying that our results cannot directly be compared with~\citet{salminen2018anatomy:2018}. Nevertheless, this benchmark serves as a possibility to verify our methodology and results. The constructed \textit{HateSpeech} dataset can be found in our released codebase.

We achieved very similar results to the \textit{IndustrySector} dataset on our public \textit{HateSpeech} dataset, as shown in Table \ref{table:public_benchmark}. The most notable difference is that for LLaMa 7B, PT outperforms PTEC. For both models, Trie Search decreases the performance of the Prompt Tuned LLM, while it slightly improves the performance for N-shot prompting of Bloom 1B7. A relevant observation made is the high standard deviation of T2T classification performance when using Bloom 1B7. This goes along with results of recent research showing that models from the Bloom family produce the most inconsistent summaries, as judged by other language models~\cite{tam2023evaluating:2023}.

\begin{table}[ht!]
\centering
\small
\setlength{\tabcolsep}{4pt}
\begin{tabular}{llrrrr}
\toprule
& \textbf{Method} & \multicolumn{2}{c}{\textbf{FLOPs}} & \multicolumn{2}{c}{\textbf{Macro F1}} \\
\cmidrule(lr){3-4} \cmidrule(lr){5-6}
& & \multicolumn{1}{c}{\textbf{Training}} & \multicolumn{1}{c}{\textbf{Inference}} & \multicolumn{1}{c}{\textbf{Mean}} & \multicolumn{1}{c}{\textbf{Std}} \\
\midrule
\multirow{8}{*}{\rotatebox{90}{Bloom 1B7}} & PTEC & 6.99e+16 & 3.96e+17 & \textbf{0.48} & 0.015 \\
& PT + TS & 8.69e+16 & 6.85e+17 & 0.233 & 0.123 \\
& PT & 8.69e+16 & 7.94e+17 & 0.318 & 0.088 \\
& CH & 6.82e+12 & \textbf{3.59e+17} & 0.063 & 0.011 \\
& KNN & 8.39e+14 & \textbf{3.59e+17} & 0.12 & 0 \\
& RadiusNN & 8.39e+14 & \textbf{3.59e+17} & 0 & 0 \\
& N-shot + TS & \textbf{0} & 2.81e+18 & 0.082 & 0.002 \\
& N-shot & \textbf{0} & 2.51e+18 & 0.055 & 0.005 \\
\addlinespace
\multirow{8}{*}{\rotatebox{90}{LLaMa 7B}} & PTEC & 1.31e+17 & 2.27e+18 & 0.437 & 0.007 \\
& PT + TS & 2.22e+17 & 2.37e+18 & 0.47 & 0.032 \\
& PT & 2.22e+17 & 3.20e+18 & \textbf{0.526} & 0.021 \\
& CH & 3.07e+13 & \textbf{1.59e+18} & 0.365 & 0.014 \\
& KNN & 3.72e+15 & \textbf{1.59e+18} & 0.195 & 0 \\
& RadiusNN & 3.72e+15 & \textbf{1.59e+18} & 0.142 & 0 \\
& N-shot + TS & \textbf{0} & 4.40e+18 & 0.094 & 0.008 \\
& N-shot & \textbf{0} & 1.16e+19 & 0.107 & 0.021 \\
\addlinespace
- & gzip & $-$ & $-$ & 0.128 & 0 \\
\bottomrule
\multicolumn{6}{l}{\footnotesize\parbox{0.95\linewidth}{gzip = Parameter-Free Classification with gzip. Other abbreviations as defined in Table \ref{table:hyperparam}.}}
\end{tabular}
\caption{Experimental results on the \textit{HateSpeech} benchmark. The method requiring the lowest FLOPs and achieving the highest macro-averaged F1 Score is highlighted in \textbf{bold} for each model. A dash ($-$) indicates that a value could not be estimated.}
\label{table:public_benchmark}
\end{table}

\end{document}